\documentclass{article} 

\usepackage{ijcai16}
\usepackage{times}
\usepackage{hyperref}
\usepackage{url}
\usepackage[pdftex]{graphicx}
\usepackage{textcomp}

\title{Performing Highly Accurate Predictions Through Convolutional Networks for\\
Actual Telecommunication Challenges}

\author{Jaime Zaratiegui, Ana Montoro \& Federico Castanedo\\
WiseAthena\\
San Francisco, CA, USA\\
\{jaime.zaratiegui,ana.montoro,fcastanedo\}@wiseathena.com}

\begin{document}

\maketitle

\begin{abstract}
We investigated how the application of deep learning, specifically the use of 
convolutional networks trained with GPUs, can help to build better predictive 
models in telecommunication business environments. 
In particular, we focus on the non-trivial problem of predicting customer churn 
in telecommunication operators. Our model, called WiseNet, consists of a 
convolutional network and a novel encoding method that transforms customer 
activity data and Call Detail Records (CDRs) into images. Experimental evaluation 
with several machine learning classifiers supports the ability of WiseNet for 
learning features when using structured input data. For this type of 
telecommunication business problems, we found that WiseNet outperforms machine 
learning models with hand-crafted features, and does not require the 
labor-intensive step of feature engineering. Furthermore, the same model has 
been applied without retraining to a different market, achieving consistent 
results. This confirms the generalization property of WiseNet and the ability 
to extract useful representations.
\end{abstract}

\section{Introduction}
Advances in the amount of available data, hardware performance and methods for training deep learning networks \cite{Hinton06} \cite{srivastava2014dropout} have shown an extraordinary predictive performance improvement in areas such as audio processing \cite{senior2013empirical}, image recognition \cite{simonyan2015very} \cite{szegedy2014going}, natural language processing \cite{mikolov2013distributed} and video analysis \cite{simonyan2014two} \cite{karpathy2014large}.


Customer churn may be defined as the loss of a customer resulting from, for example, the customer switching to a competitor's product or service. Predicting churn of prepaid telecommunication customers is a complex task since there is no specific termination date of their contract.
In this circumstances, deactivation occurs after a period of time during which the customer did not perform a balance replenishment. It can be further observed that, once the customer has decided    or is in the process of deciding whether to transition, the customer's phone usage patterns
may start to change. The sooner these changing patterns are detected, the more opportunities and time are available to the telecommunications service provider to try to retain the customer. Being able to predict which customers most likely stop their replenishment gives companies a competitive advantage to retain customers since they can perform specific commercial actions to reduce the likelihood that the customer will churn. This problem has been approached by training binary classifiers with the behavior of known churners/non-churners and then applying the trained model to predict future behaviors. Previous state-of-the-art works tackled the problem by training machine learning models with hand-crafted features \cite{huang2012customer} \cite{DBLP:journals/corr/LucianoRL15} \cite{DBLP:journals/corr/WilsonKMKSC15}
\cite{yoon2010prediction} \cite{burez2009} or by applying social network analysis \cite{richter2010predicting}.

Whereas the superiority of ConvNets against traditional machine learning models is clear when dealing with noisy unstructured data (e.g.~images), their application to structured input data has been less explored. In this article, we show that the application of ConvNets using structured data outperforms traditional supervised classifiers with hand-crafted features, with no need to perform the labor-intensive step of feature engineering. To the best of our knowledge, this is the first work which provides an exhaustive comparison between feature-engineering machine learning models, and ConvNets for a real telecommunication business problem.

Our main contributions are the following:
\begin{itemize}
 \item The development of a novel method to encode customer behavior into images specifically crafted for applying
       structured data from a real telecommunication business problem to WiseNet.
 \item An experimental evaluation with different machine learning models showing the ability of WiseNet
       to learn features when dealing with structured data.
  \item A comparison with a production system developed using hand-crafted features showing the advantage of WiseNet in this problem.
  \item The application of the same network architecture, weights and hyperparameters to a different market without retraining, showing
  the generalization and transfer learning property of the WiseNet model.
\end{itemize}

In the next section we provide details about the type of data we deal with and the approach to encode customer behavior into images. Then the WiseNet network architecture employed to solve this problem is discussed (Section \ref{network}) and experimental results using WiseNet and machine learning classifiers over the same datasets are given in Section \ref{exp}. Finally, the conclusions of this work are shown in Section \ref{summary}.

\section{Data representation}\label{data}
In a telecommunication environment, several data sources are available from the customer behavior point of view, such as Call Detail Records (CDRs), balance replenishment events (topups), geographical activity, social network activity, data-usage behavior, promotions/marketing campaigns, to name a few.
 In order to come up with a general model our goal is
 to avoid specific details and therefore 
preventing domain specific representations as much as possible. 
 In this work, we only use the following data from the customer behavior:
\begin{itemize}
  \item Call Detail Records (CDRs): providing log information with details about each call made by the customer. This information
  includes the cell tower where the call was made, the number originating the call\footnote{To enforce security and privacy issues the 
  destination and originating number are always encrypted.}, when the call was made (time-stamp), 
  the duration of the call, the destination number, and the service identification (incoming/outgoing call or sms).
  \item Topups: describing a history of balance replenishment events for each customer, basically the customer-id, the time-stamp and the amount
  spent.
 \end{itemize}

We were motivated to employ these pieces of data seen the impressive results of ConvNets, from among the myriad of deep learning architectures. The basics of ConvNets are template layers which are learned during training on the image pixels. Therefore, we need to encode customer behavior data into artificial images. This encoding step can be seen as a transformation from the input space to a specific training domain, similarly to what is done with word representations in the word2vec model \cite{mikolov2013efficient}.

\subsection{Image encoding}
In our work, customer data is encoded into a linear time representation. Hence, time is linearly mapped into the $x$-axis of the image. The $y$-axis is reserved for each one of the data sources, i.e.~outgoing call activity (MOC), incoming calls (MTC) and topups.

The image width spans the whole period of time we considered in the training stage, which is exactly four weeks or 28 continuous days of user activity. Furthermore, we have split each day into 12 equally sized time slices and, as a result, each of these slices, i.e.~pixels, spans a period of two hours. Therefore, the total image width and height in pixels is $336\times3$.

\begin{figure}[ht]
\begin{center}
\includegraphics[width=2cm,height=1cm]{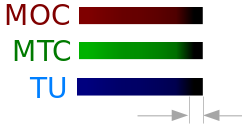}
\end{center}
\caption{Different pixel categories and intensity range in the RGB scale for each one of the customer events. The size of a 2-hour pixel is delimited by the gray arrows. The vertical arrangement of the three data categories MOC/MTC/topups is kept equal in all images.}
\label{fig:pixel-intensities}
\end{figure}

The value or intensity of each pixel in the image is proportional to the customer's activity in each of the three possible categories. For example, let us assume a customer has a 20 minute outgoing call in a certain time slice $t_n$, the base value for that pixel would then be $20/120=1/6$, as 120 minutes is the maximum activity that can be registered. Due to the fact that most variance is found closest to zero activity, we scale the pixel's base value $I$ according to the following power law
\begin{equation}
I=I^\alpha,
\label{eq:intensity}
\end{equation}
where $\alpha$ is chosen to be equal to $1/7$ 
in order to better exploit variance in those pixels that have the lowest base values. Continuing with the example of a 20 minute outgoing call, we would finally have that the normalized value of the pixel will be $I=6^{-1/7}=0.774$, which in an 8-bit image corresponds to the pixel value 197. We have chosen each of the RGB channels to represent separately each of the three categories MOC (red), MTC (green) and topups (blue), as shown in Figure \ref{fig:pixel-intensities}.

For the topup activity channel, we have chosen a different way of representing the input data. In this case, the balance topup activity is predominantly done by redeeming coupons of discrete value and seldom by doing a bank transfer of a non-preset quantity. Consequently, the pixel values will be mostly discrete. In this case, the pixel intensity will be linearly proportional to the amount replenished but will saturate at a certain point. We have set as saturation value the maximum value of a single prepaid coupon which is available for purchase. Of course, a single customer could redeem several of these maximum face value coupons in the same two hour period but the value of that pixel in the blue channel will not be affected as it will already be saturated.

\begin{figure}[ht]
\begin{center}
\includegraphics[width=8.5cm]{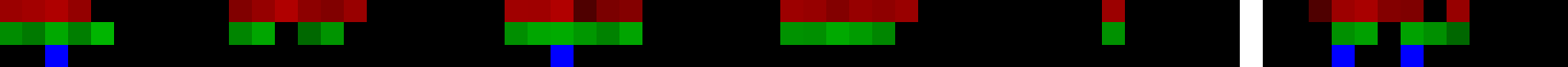}
\end{center}
\caption{An encoding example of a six day period for a single customer with a high degree of activity in the three categories. Start of the week white mark can be observed in the right part of the image. Time increases from left to right.}
\label{fig:single-customer}
\end{figure}

A single extra feature has been added to the images, that is a white $1\times3$ pixel mark at each point where local time is Monday at 00:00 hours. This is done at the expense of all activity data at that point as the white vertical stripe deletes any previous value present at that point. The purpose of this stripe is identifying the weekly periodicity inherent to calendar days besides any other recurring activity that the user could have (see Figure \ref{fig:single-customer}).

Each of the images is labeled according to the user topup activity in the 28 consecutive days following to the training data shown in the image. Churners are defined as those who did not have any topup activity in this period and therefore, their corresponding images are labeled as 1.

In order to train and validate the models we followed a standard train, test and validation data partitioning. We took a sample of 132779 different customers in a three-month period from January 2015 till end of March 2015. The four week training period is taken at random within the available time span, given the condition that there is enough data for the additional 28 days required for labeling the image. The train set was composed of 102K customers, validation set of 18K and test set of 37K different customer instances. The validation set was used to tune the classifiers' optimal parameters such as number of neurons, depth of the hidden layers or learning rate. The performance metrics of the classifier such as area under the curve (AUC) or logarithmic loss, are evaluated using the test set.

\section{Network architecture}\label{network}
As we have mentioned before, to develop the WiseNet architecure, we selected ConvNets due to their ability to generalize to new samples and to perform feature learning. ConvNets have a specialized connectivity structure which exploits the strong spatial correlation exhibited in the customer-behavior artificial images.
Another reason for using convolutions is that our patterns may be shifted around and it would be desirable to detect this based on the position in the image.

Once features are learned using ConvNets, they are passed to a fully connected neural network which combines them in order to classify the input image into one specific class. In our case, images represent customers' behavior and the target class indicates if they are active or inactive customers (churners).

We used pooling layers and experimented with different network architectures, activation functions and initialization procedures. We found that networks are remarkably sensitive to the initialization and the activation functions.

ConvNets with pooling layers have some translation invariance built into their architecture. A pooling layer summarizes the information of the convolution units in a smaller number of units. A unit in a pooling layer receives input from a small neighborhood of convolution units in the layer below. Pooling layers can be divided into sum, average or max pooling. We use max-pooling layers, which means that the strongest convolutional filter response in the small region is passed on to the next layer. 




Regarding activation functions, we used parametric rectifier linear units (PReLU) \cite{he2015delving}. This type of activation function simplifies backpropagation, makes the learning phase faster and avoids saturation issues.

The design of the architecture of the network can be split into two stages or modules. In the first stage, convolutional and spatial pooling layers are applied sequentially taking as inputs the set of RGB images. In the second stage, a series of fully connected hidden layers are used with the objective of exploiting the features extracted in the first stage.

\begin{figure}[!ht]
\begin{center}
\includegraphics[width=8.5cm]{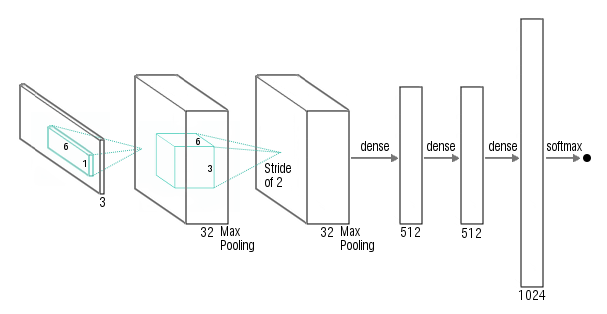}
\end{center}
\caption{Schematic representation of WiseNet's architecture shown in Table~\ref{table:netarch}. The convolution kernels applied in the first stage are shown colored in cyan.}
\label{fig:netarch}
\end{figure}

The convolution filters used in the ConvNet are relatively small. In the first hidden layer, we have used a rectangular $6\times1$ kernel with a stride of 1. In this step, using a 1-pixel high filter, we aim to extract features which are dependent only on each of the rows of the images, i.e.~the three different activity categories will not be mixed in this operation. After the second convolution layer, the data from the three rows of the image will be mixed as we are using a $6\times3$ filter. The width of the convolution layers is also quite small, just 32 channels wide. The exact specifications of the WiseNet are shown in Table~\ref{table:netarch} and a visual illustration of the architecture is represented in Figure~\ref{fig:netarch}.

\begin{table}[!ht]
\begin{center}
\begin{tabular}{lcccc}
  \hline
  Layer & dim & channels & kernel & stride \\
  \hline \hline
  Input RGB & 336 $\times$ 3 & 3 &  & \\
  \\
  Convolutional & 331 $\times$ 3 & 32 &  6 $\times$ 1 & 1 \\
  Max Pooling & 326 $\times$ 3 & 32 &  6 $\times$ 1 & 1 \\
  Convolutional & 321 $\times$ 1 & 32 &  6 $\times$ 3 & 1 \\
  Max Pooling & 161 $\times$ 1 & 32 &  2 $\times$ 1 & 2 \\
  \\
  FC-512 & & & & \\
  FC-512 & & & & \\
  FC-1024 & & & & \\
  FC-2 & & & & \\
  soft-max \\
  \hline
\end{tabular}
\end{center}
\caption{The architecture of WiseNet can be divided into two stages. The first one, in which an alternating combination of convolutional and max pooling layers are applied in order to extract low-level features. The second stage consists of three fully connected layers ending in a soft-max activation function.}
\label{table:netarch}
\end{table}

\section{Experimental results}\label{exp}


In this section, we present the performance metrics of the image classification applied on the test dataset described in Section~\ref{network}. For the evaluation and comparison of models, we have used the following four different metrics with decreasing order of importance: The Area Under the Curve (AUC); The logarithmic likelihood of the classification (logloss); The true positive fraction in the top-5\% quantile (TP5) and the Brier score.

Whereas TP5 metric can be considered as one of the best metrics if we were to use the outcome of the model to optimize the detection and prevention of customer churn, we prioritize AUC as our main indicator, the reason being considered a more general metric because it takes into account not only the highest probabilities but all possible cutoffs.

Although the metrics used for performance evaluation of the models are the ones described previously, to select the best model in the training phase we took the one with the minimum log-loss in the validation dataset. By following this strategy, we expect to choose the least biased model and therefore the best one at generalizing. We show in Table~\ref{table:results} the WiseNet error, with a cutoff probability of 0.5, and log-loss of the best selected architecture evaluated in the three datasets: train, test and validation.

\begin{table}[!ht]
\begin{center}
\begin{tabular}{rcc}
  \hline
  Dataset & Error & Log-loss \\
  \hline \hline
  train & 0.1967 & 0.4330 \\
  validation & 0.1995 & 0.4273 \\
  test & 0.1979 & 0.4274 \\
  \hline
\end{tabular}
\end{center}
\caption{Selected metrics of the best model on train, validation and test datasets. The small difference between these numbers indicates that the WiseNet model is performing with very little overfitting.}
\label{table:results}
\end{table}

Besides the metrics shown in Table~\ref{table:results}, the performance on the test dataset of our main indicator (the AUC) is 0.8787, which can be considered generally as a good result for a binary classifier. The values of the remaining metrics are shown in Table~\ref{table:resultscomparison}. One remarkable aspect of the results shown in Table~\ref{table:results} is the really small difference in the log-loss between test and validation datasets. Although the difference is also small between the train set and the remaining two, it can be due to the fact that test and validation sets contain the same fraction of instances per class, whereas the training dataset is constructed as a balanced set. We also realized that the extensive use of dropout regularization layers had the effect of minimizing the overfitting of the model and improving generalization.

The framework we used for implementing the convolutional method is the publicly available cxxnet. Although this tool can be used natively as a distributed multi-GPU toolkit, due to the size of the problem we have not considered using multiple GPUs or a distributed GPU system at this moment. Therefore, our training stage was done in a single GPU machine equipped with one NVIDIA K20. A single manipulation is done by cxxnet previous to the training stage, the average image of the training dataset is calculated and then the value of every individual pixel in the image is subtracted to the mean.

\subsection{Comparison with other models}

In order to compare our best WiseNet model with other machine learning techniques we used the same artificial images as comma-separated value (CSV) files. The information contained in the RGB channels was flattened into a single value per pixel and stored into CSV file. To maintain the information at the start-of-the-week mark, we added a new variable in the CSV which fills in this gap. This variable indicates the offset, in pixels, at which the random 4-week sample has been cropped. As a result, each customer was described by a feature vector of 1009 dimensions plus one extra column containing the class label.

We considered the performance of four well-known machine learning algorithms compared to the best performing WiseNet model: randomForests, generalized linear models (GLM), generalized boosted machines (GBM) and extreme gradient boosting (xgboost). We show on Table~\ref{table:resultscomparison} the performance comparison for all models ranked by AUC in decreasing order from top to bottom. The optimal set of hyper-parameters was determined using the validation dataset and optimizing the log-loss.

\begin{table}[!ht]
\begin{center}
\begin{tabular}{rcccc}
  \hline
  & AUC & Log-loss & TP5 & Brier \\
  \hline \hline
  WiseNet & 0.8787 & 0.4274 & 0.8929 & 0.1383 \\
  xgboost & 0.8561 & 0.4722 & 0.8908 & 0.1594 \\
  GBM & 0.8512 & 0.4995 & 0.8750 & 0.1662 \\
  GLM & 0.8228 & 0.6782 & 0.7592 & 0.2433 \\
  randomForest & 0.8169 & 1.2482 & 0.8636 & 0.2018 \\
  \hline
\end{tabular}
\end{center}
\caption{Test metrics of the best model of each algorithm class. It can be noticed that WiseNets outperform all the other models in every indicator considered.}
\label{table:resultscomparison}
\end{table}

\begin{figure}[ht]
\begin{center}
\includegraphics[width=8.5cm]{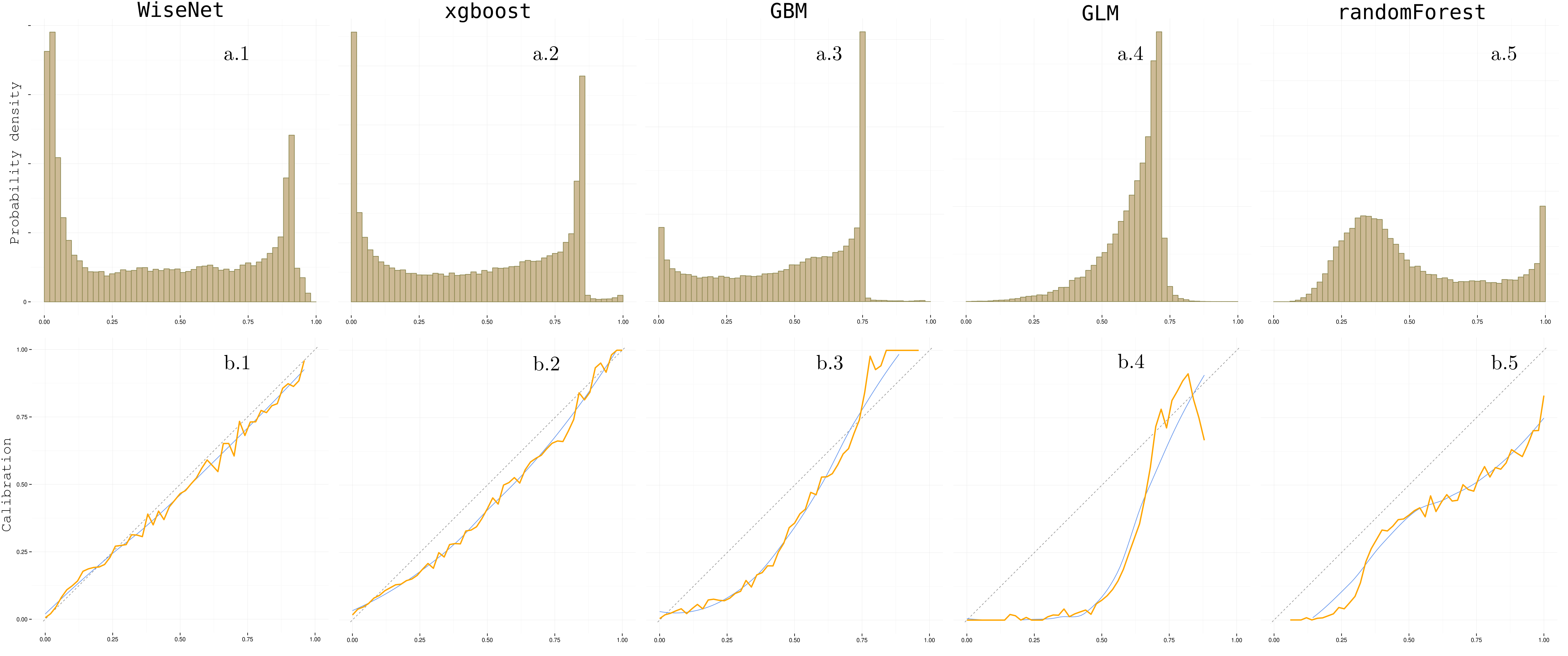}
\end{center}
\caption{Unnormalized probability density distributions and calibration curves of the models studied in this section. For the calibration curve, the orange curve represents the measured true positive rate for each probability step. A perfectly calibrated model would be represented as a straight line with a slope of 1. The blue curve is plotted as a guide and is just a smoothed version of the orange one.}
\label{fig:modeloutputs}
\end{figure}

Looking at the results on Table~\ref{table:resultscomparison}, our WiseNet outperforms all the other models in every metric studied. Therefore, we can deduce that, under the conditions here reviewed, WiseNets can extract a higher amount of information and learn specific feature representations. It can be noticed the relatively poor performance of the randomForest model, specially under the log-loss metric, because this model is optimized for minimizing misclassification error instead of log-loss. This particular model outputs a higher density of predictions in the top probability range $p>0.9$ when compared with other methods. This characteristic combined with the fact that its output is not particularly well calibrated in that probability range, as can be seen in Figure~\ref{fig:modeloutputs} b.5, damages the performance of its logarithmic loss score.

\subsection{Comparison with feature-engineering models}

Due to the lack of published literature about machine learning techniques applied to churn prediction in prepaid telecommunication networks, we present in this section a performance comparison between one of our in-house models which makes extensive use of feature engineering and a WiseNet. These features, developed and fine tuned during months of work, condense our knowledge of this particular market and its situation.

The in-house model has been trained with the same set of customers as previous models. It even includes new sources of data not used in the WiseNet model, such as the geographical location of customers, use of other mobile services, categorical variables and social relationships with other customers. Another significant difference with the WiseNet model is that the training dataset size of the feature-engineering model is larger but the same test dataset has been used nonetheless.

\begin{table}[!ht]
\begin{center}
\begin{tabular}{rcccc}
  \hline
  & AUC & Log-loss & TP5 & Brier \\
  \hline \hline
  WiseNet & 0.8787 & 0.4274 & 0.8929 & 0.1383 \\
  F-Model & 0.8552 & 0.4602 & 0.7184 & 0.1528 \\
  \hline
\end{tabular}
\end{center}
\caption{Metrics comparison on the test dataset between the WiseNet and the model using features (F-Model).}
\label{table:resultscomparisonfmodel}
\end{table}

As we can see in Table~\ref{table:resultscomparisonfmodel}, our WiseNet model outperforms the one using feature engineering in all the metrics considered, and this difference is particularly remarkable under the TP5 metric.

\subsubsection{Comparison with external models for churn prediction}

There are several comprehensive reviews of the accuracy of customer churn predictive
models in the telecommunication industry~\cite{neslin2006} or in banking, retail and even insurance~\cite{klepac2014}.
Although the studied models do not target the same specific markets neither they have been measured using the same performance
metrics as we have, we can compare the results that appear in~\cite{neslin2006}
using their definition for the top-decile lift. In the best performing case studied by~\cite{neslin2006}
the top-decile lift was $3.01$, below the score of $3.52$ that WiseNet achieves and far beyond the reported average of 2 units.

\cite{burez2009} reports a better lift performance of 4.4099 for a Logistic regression classifier applied to churn detection on a Mobile market
 but it underperforms WiseNet in AUC as it only scores 0.6785. Using text mining techniques, \cite{coussement2008} developed a classifier for
predicting customer churn with a top-decile lift performance of 3.02 and an AUC of 0.7775, both figures underperform WiseNet metrics.

\subsection{Application in other markets}

\begin{figure}[htb]
\begin{center}
\includegraphics[width=8cm]{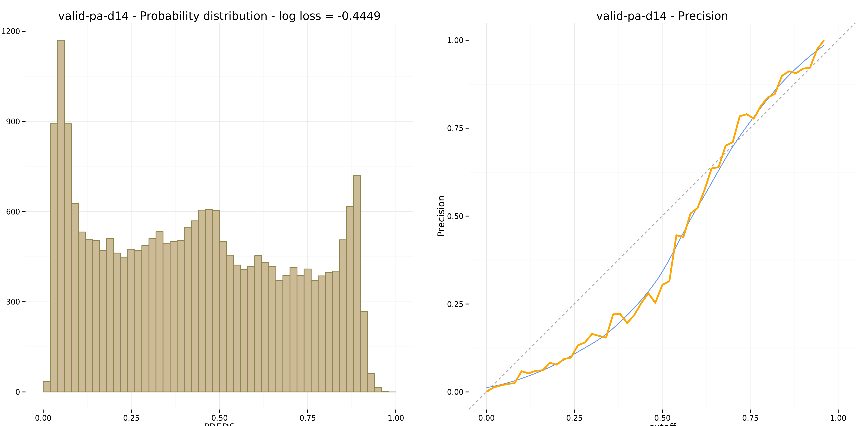}
\end{center}
\caption{Unnormalized probability density distributions and calibration curves of the WiseNet model applied to a different market without re-training (using the same network architecture, weights and hyperparameters).}
\label{fig:marktet2}
\end{figure}

It can be conjectured that similar results can be achieved using the pre-trained WiseNet model in other markets, influenced by different
factors. Inspired by this hypothesis, we evaluated the best WiseNet using data from other country and without re-training of the model, therefore using
the same network architecture, weights and hyperparameters.
The goal was to test the transferable property or generalization error of the model. 
We were surprised by the extremely good results obtained with this model (see Table \ref{table:results2markets}).
Two different reasons may justify this salient characteristic: (i) the customers' behavior is similar in both countries and/or (ii) the model
is capable of extracting general patterns which can be applied across different markets. The obtained unnormalized probability density distributions and calibration curves
are depicted in Figure \ref{fig:marktet2}. The effect of predicting without re-training can be appreciated in the calibration graph.

\begin{table}[!ht]
\begin{center}
\begin{tabular}{ccccc}
  \hline
  WiseNet & AUC & Log-loss & TP5 & Brier \\
  \hline \hline
  market-1 & 0.8787 & 0.4274 & 0.8929 & 0.1383 \\
  market-2 & 0.8788 & 0.4449 & 0.9163 & 0.1428 \\
  \hline
\end{tabular}
\end{center}
\caption{Metrics comparison on the test datasets of two different markets. WiseNet model was learned using the train dataset of market 1 and tested using test datasets of two different markets 1 and 2.}
\label{table:results2markets}
\end{table}

\subsection{Learned representations analysis with t-SNE embeddings}

\begin{figure}[!htb]
\begin{center}
\includegraphics[width=8.8cm]{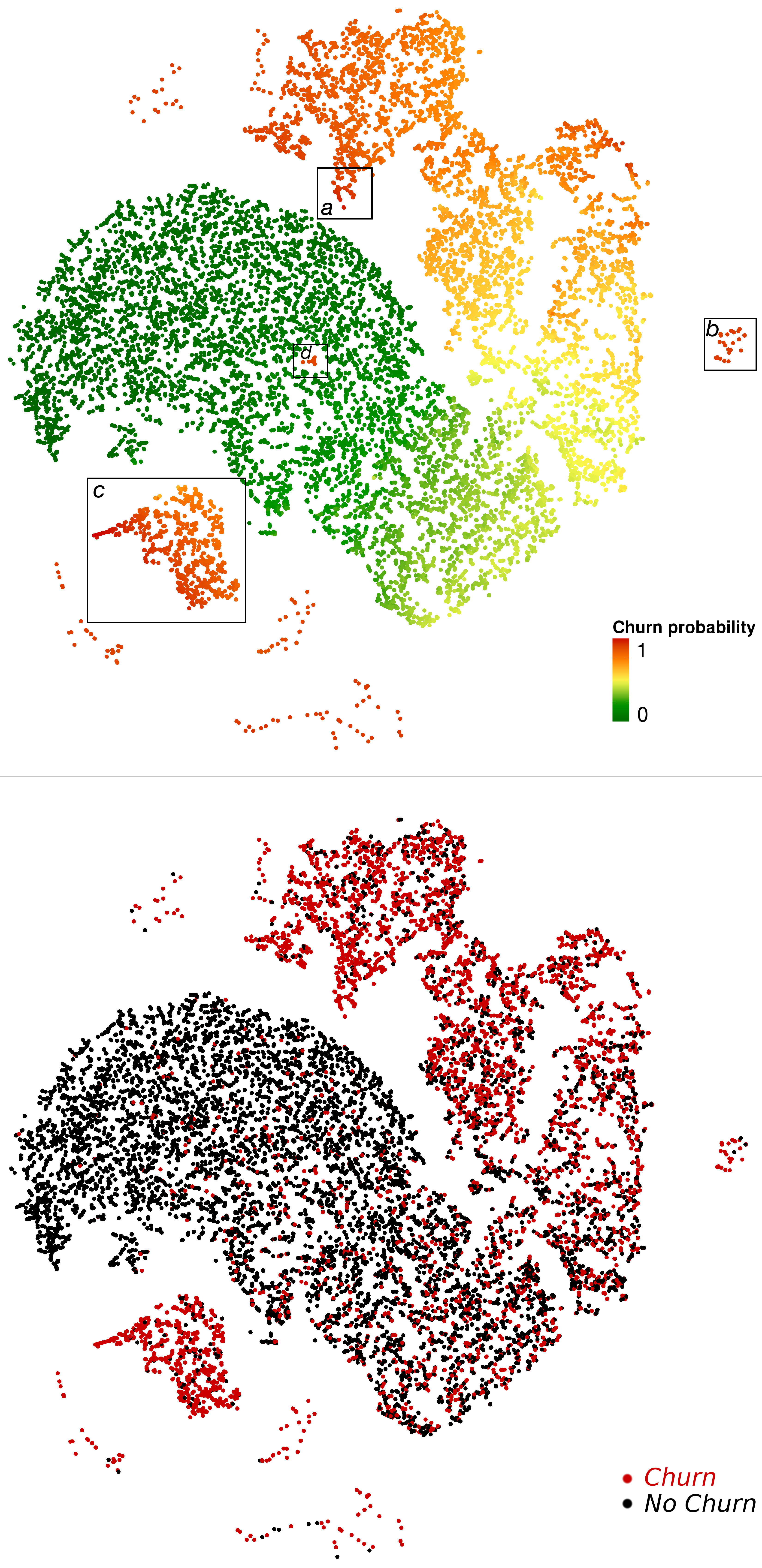}
\end{center}
\caption{Two dimensional t-SNE embedding of the test dataset using the states of the 1024 neurons of the last hidden layer. On the top image, each point is colored according to the probability given by WiseNet for each user belonging to the churner class, red corresponds to the highest probability and green the lowest. On the bottom image the color corresponds to the observed class with red being a churner.}
\label{fig:tsne}
\end{figure}

In order to analyze the possible configurations associated with churn, we have applied a t-Distributed Stochastic Neighbor Embedding (t-SNE) to perform a dimensionality reduction on a subsample of 25 thousand users from the test dataset. The input variables passed to the t-SNE are the states of the neurons in the last FC-1024 hidden layer. The results of the two-dimensional embedding can be seen in Figure~\ref{fig:tsne}.

On the top part of Figure~\ref{fig:tsne}, we can see the t-SNE embedding in which we have colored each point according to its probability of belonging to the churner class. It can be noted that despite not including this probability among the inputs passed to the dimensionality reduction, there is a smooth distribution along the map of this variable. In particular, in the largest cluster of points, the churn probability seems to change almost continuously from one extreme to the other. The only exception is the one being delimited within the small square $d$.

Some of the most remarkable aspects of the top part of Figure~\ref{fig:tsne} have been highlighted with labels $a$ to $d$. As mentioned before, the probability across the largest cluster seems to change smoothly from the left green part to regions like $a$ in which there is a high probability of churn. As this change is continuous, we can assume that the difference between these points is just quantitaive, i.e.~total call time or total amount of topups diminishes, but there is not a real qualitative change in phone activity usage. However, clusters like $b$ and $c$ are noticeably separated from the main cluster so there is a higher chance that they show a qualitative difference, rather that just a quantitative one, between the largest cluster. The location of the high-churn probability cluster $d$, completely embedded in a region of low churn probability, is quite a surprise. It could be due just to a simple convergence issue of the t-SNE algorithm or its location could have deeper foundations. Users in cluster $d$ share an activity pattern similar to their non-churner neighbors but instead, they are labeled with a very high churn probability.  More analysis would be needed in order to understand this situation as it could lead to a localized anomaly or to the discovery of a new activity pattern.

Another singular feature of the t-SNE embeddings is that there are no isolated clusters with low-churn probability, other than the largest one. We could interpret this as different causes for customers to churn from the service provider e.g., bad network coverage, expensive costs$\ldots$ etc.~but the reason to stay can be summarized as a simple cause, e.g. the customer needs and wants the service.

\section{Summary and conclusions}\label{summary}

Nowadays, machine learning engineers spent most of their time finding good representations for their models and a huge research effort is
done trying to come up with better features.

In this work we investigated the application of ConvNets trained with GPUs to build better predictive models in telecommunication business
environments and developed the WiseNet model. For the  particular problem that we considered, it has been shown that WiseNet outperforms commonly used machine
learning models using the same input data and effectively discover features when there are some topological structure
in the data (temporal or spatial). We have also shown that it is possible to automatically learn the representations required
for a specific business problem, providing more accurate predictions than traditional machine learning models built with hand-crafted features.
This finding is translated into an exponential decrease in the required human-effort for feature engineering compared to traditional machine learning models.

Furthermore, similar results were achieved using the pretrained WiseNet model in other markets, which may be influenced by different factors. So, we
confirmed the generalization property of the WiseNet model and its ability to discover useful representations. As an ongoing work we are applying
the WiseNet model to other prediction tasks with promising preliminary results.

We have also experimented with an extended data representation and network model, however due to the confidentiality restrictions we
are not allowed to disclose them.

\clearpage

\bibliographystyle{named}
\bibliography{WiseNet}

\end{document}